\documentclass[conference]{IEEETran}
\usepackage[utf8]{inputenc} 
\usepackage[T1]{fontenc}    
\usepackage{hyperref}       
\usepackage{url}            
\usepackage{booktabs}       
\usepackage{amsfonts}       
\usepackage{nicefrac}       
\usepackage{microtype}      
\usepackage{xcolor}         
\usepackage{amsmath,amssymb}
\usepackage{wrapfig}
\usepackage{stfloats}
\usepackage{graphicx}
\usepackage{float}
\usepackage{caption}
\usepackage{subcaption}
\usepackage{bm}
\usepackage{listings}
\usepackage{multicol}
\usepackage{multirow}

\NewDocumentCommand{\codeword}{v}{%
\texttt{\textcolor{blue}{#1}}%
}

\title{CaAdam: Improving \textit{Adam} optimizer using connection aware methods }

\author{%
  \IEEEauthorblockN{%
    Rémi Genet\IEEEauthorrefmark{1}\textsuperscript{\textsection} and
    Hugo Inzirillo\IEEEauthorrefmark{2}\textsuperscript{\textsection}
  }%
  \IEEEauthorblockA{\IEEEauthorrefmark{1} DRM, Université Paris Dauphine - PSL}%
  \IEEEauthorblockA{\IEEEauthorrefmark{2} CREST, Institut Polytechnique de Paris}%
}

\begin{document}
\thispagestyle{plain}
\pagestyle{plain}
\maketitle
\begingroup\renewcommand\thefootnote{\textsection}
\footnotetext{These authors contributed equally. Author ordering determined by odd or even days.}
\endgroup

\begin{abstract}
\begin{quote}
We introduce a new method inspired by \textit{Adam} that enhances convergence speed and achieves better loss function minima. Traditional optimizers, including Adam, apply uniform or globally adjusted learning rates across neural networks without considering their architectural specifics. This architecture-agnostic approach is deeply embedded in most deep learning frameworks, where optimizers are implemented as standalone modules without direct access to the network's structural information. For instance, in popular frameworks like Keras or PyTorch, optimizers operate solely on gradients and parameters, without knowledge of layer connectivity or network topology. Our algorithm, \textit{CaAdam}, explores this overlooked area by introducing connection-aware optimization through carefully designed proxies of architectural information. We propose multiple scaling methodologies that dynamically adjust learning rates based on easily accessible structural properties such as layer depth, connection counts, and gradient distributions. This approach enables more granular optimization while working within the constraints of current deep learning frameworks. Empirical evaluations on standard datasets (e.g., CIFAR-10, Fashion MNIST) show that our method consistently achieves faster convergence and higher accuracy compared to standard \textit{Adam} optimizer, demonstrating the potential benefits of incorporating architectural awareness in optimization strategies.
\end{quote}
\end{abstract}

\section{Introduction}
In deep learning, optimizers play a crucial role in the efficient training of neural networks. Each of the best-known deep learning libraries (\href{http://keras.io/optimizers}{Keras\cite{chollet2015keras}},\href{https://pytorch.org/docs/stable/optim.html}{Pytorch\cite{paszke2019pytorch}}) has implemented different algorithms to optimize gradient descent. Among the most popular optimizers, Adam (Adaptive Moment Estimation) \cite{kingma2014adam} is widely used for its ability to dynamically adjust learning rates according to the first- and second-order moments of the gradients. However, Adam applies a global uniform learning rate to all parameters which impacts the convergence speed. This weakness is all the more noticeable for deep, complex neural networks, where different layers may require different update rate adjustments.

\medskip

This work aims to challenge the traditional paradigm of treating neural network optimization as a uniform process across all layers. We propose CaAdam (Connection Aware Adam), an extension of the Adam optimizer that introduces adaptive scaling strategies based on the structural properties of neural networks. Unlike Adam, which adjusts parameter updates globally, our approach allows learning rates to be adapted more granularly, considering the number of connections per layer, network depth and gradient distribution. This flexibility allows CaAdam to increase the speed of learning within layers where faster update are needed and stabilize training for layers where more gradual updates are beneficial. While our current implementation focuses on basic metrics such as connection counts and layer depth, we view this as an initial exploration into architecture-aware optimization. These simple proxies for network structure demonstrate significant improvements in training dynamics, suggesting a rich space for future research. More sophisticated approaches might consider topological features, layer interactions, or dynamic architectural characteristics during training. We believe this work opens a new perspective on how optimization strategies could be better aligned with neural network architectures, potentially leading to more efficient and effective training methodologies. Our codes are available at \href{https://github.com/remigenet/Caadam}{CaAdam repository} and can be installed using the following command: \codeword{pip install CaAdam}.
\section{Related work}
Gradient descent, a foundational optimization algorithm in machine learning, traces its origins to Cauchy's method of steepest descent introduced in 1847 \cite{cauchy1847methode}. This mathematical foundation laid dormant for over a century before finding renewed purpose in the realm of computational optimization. The method gained significant traction in the 1960s through the work of Davidon, Fletcher, and Powell, who developed practical applications for numerical optimization \cite{fletcher1963rapidly}.
When it comes to deep learning, gradient descent has emerged as an indispensable cornerstone \cite{du2019gradient,andrychowicz2016learning}. The algorithm's ability to handle high-dimensional optimization problems made it particularly well-suited for neural networks, as demonstrated in seminal work by Rumelhart et al. \cite{rumelhart1986learning}. Several methods have since been introduced to enhance its effectiveness \cite{ruder2016overview}. In the context of deep neural networks \cite{lecun2015deep}, where models are complex and have thousands or even millions of parameters, this optimization is crucial if the model is to learn from the data.
Gradient descent objective is to minimize $f(\theta)$, an objective function parametrized by $\theta \in \mathbb{R}^d$. The procedure consist of updating $\theta$ in the opposite direction of $\nabla_\theta L\theta;x_i,y_i$. This update step is performed using a learning rate, denoted $\eta$. Taking this simple explanation, we can write the procedure of updating $\theta$ such:
\begin{equation}
    \theta = \theta - \eta \cdot \nabla_\theta L(\theta;x^{(i)},y^{(i)})
    \label{eq:sgd}
\end{equation}
where $x_i = (x_{i,1},x_{i,2},...,x_{i,m})$ represents the feature vector for the i-th sample, containing $m$ features of the dataset $D \triangleq {(x_i,y_i)|i = 1,...,N}$. $y_i$ is the target value of the i-th sample. In some problems $y_i$ may be known (supervised learning) or unknown (unsupervised). By adjusting model parameters towards a minimum of the loss function $L(\theta)$; 

\begin{equation}
    L(\theta) = \frac{1}{N}\sum_{i=1}^{N} l(f_{\theta}(x^{(i)}),y^{(i)})
\end{equation}
where $l(.)$ is a specific loss function (e.g mean squared error, cross entropy,etc.) The gradient descent helps neural networks generate accurate predictions and generalize efficiently to new data. Stochastic gradient descent (SGD) performs parameter update for each data point Eq.\eqref{eq:sgd}, this configuration allow to jump to new local minima easily compared to the Batch gradient which apply the update of the gradient over the entire dataset. Mini-batch gradient descent finally takes the best of both worlds \cite{ruder2016overview} perform an update on each mini-batch of training sample;
\begin{equation}
    \theta = \theta - \eta \cdot \nabla_\theta \cdot L(\theta;x^{(i:i+b_s)},y^{(i:i+b_s)}),
\end{equation}
where $b_s$ denotes the mini-batch size. Various optimization algorithms have been proposed to improve the convergence and stability of gradient descent. Here, we discuss key advancements, beginning with Adagrad and leading to later refinements like Adam and Nadam, each building on and addressing limitations of prior methods.

\cite{duchi2011adaptive} introduced \textit{Adagrad} an adaptive gradient descent algorithm that modifies the learning rate individually for each parameter. The update is based on the historical accumulation gradients. Let us define $\theta={\theta_1,\theta_2,...,\theta_j}$ the vector of parameters. For each parameter $ \theta_i, i\in {1,2,...,j}$ at timestep $t$ Adagrad computes $g_{t,i}$

\begin{equation}
    g_{t,i} = \nabla \theta_t L(\theta_{t,i})
\end{equation}

to provide the following update:
\begin{equation}
    \theta_{t+1,i} = \theta_{t,i} - \eta \cdot g_{t,i}
    \label{eq:adagrad_1}
\end{equation}

However, Adagrad modifies this by scaling the learning rate based on the sum of squared gradients up to the current time step:

\begin{equation}
    \theta_{t+1, i} = \theta_{t, i} - \frac{\eta}{\sqrt{G_{t, ii} + \epsilon}} \cdot g_{t, i}
    \label{eq:adagrad_2}
\end{equation}
where \( G_{t} \) is a diagonal matrix with elements \( G_{t, ii} = \sum_{\tau=1}^t g_{\tau, i}^2 \), and \( \epsilon \) is a small constant for numerical stability. This adjustment enables larger updates for infrequent parameters, making Adagrad effective for sparse data \cite{dean2012large}. Building upon $Adagrad$, \cite{zeiler2012adadelta} proposed \textit{Adelta}.
Instead of accumulating all past squared gradients, Adadelta uses a decaying average of gradients, updating the squared gradients recursively as follows:
\begin{equation}
    E[g^2]_t = \gamma E[g^2]_{t-1} + (1 - \gamma) g_t^2
\end{equation}

where $\gamma \approx 0.9$ is similar to the momentum term. This allows Adadelta to maintain a more consistent learning rate, avoiding the gradual reduction to zero seen in Adagrad. The parameter update rule is given by:

\begin{equation}
    \Delta \theta_t = - \frac{\eta}{\sqrt{E[g^2]_t + \epsilon}} g_{t}
\end{equation}
where $E[g^2]_t$ replaces $G_{t} $ from Adagrad \eqref{eq:adagrad_2}. This approach allows to remove the need for a fixed learning rate. Proposed independently around the same time, RMSprop introduced by Geoff Hinton in his lecture notes\footnote{\url{http://www.cs.toronto.edu/~tijmen/csc321/slides/lecture_slides_lec6.pdf}} also seeks to mitigate Adagrad’s decaying learning rate by using an exponentially decaying average of squared gradients:

\begin{equation}
    E[g^2]_t = 0.9 E[g^2]_{t-1} + 0.1 g_t^2
\end{equation}

The RMSprop update rule is:

\begin{equation}
        \theta_{t+1} = \theta_{t} - \frac{\eta}{\sqrt{E[g^2]_t + \epsilon}} g_{t}
\end{equation}
This update method has been popularized in deep learning applications due to its stability with online and non-stationary data.

Adam (Adaptive Moment Estimation) \cite{Kingma2015} introduces both an exponentially decaying average of past gradients \( m_t \) and squared gradients \( v_t \):

\[
\begin{aligned}
m_t &= \beta_1 m_{t-1} + (1 - \beta_1) g_t \\
v_t &= \beta_2 v_{t-1} + (1 - \beta_2) g_t^2
\end{aligned}
\]

These averages are then bias-corrected as:

\[
\begin{aligned}
\hat{m}_t &= \frac{m_t}{1 - \beta^t_1} \\
\hat{v}_t &= \frac{v_t}{1 - \beta^t_2}
\end{aligned}
\]

The Adam update rule is then:

\[
\theta_{t+1} = \theta_{t} - \frac{\eta}{\sqrt{\hat{v}_t} + \epsilon} \hat{m}_t
\]

where default values \( \beta_1 = 0.9 \), \( \beta_2 = 0.999 \), and \( \epsilon = 10^{-8} \) are commonly used. Adam's bias correction and adaptive learning rates have made it a highly effective optimization algorithm in practice.

\subsection{AdaMax}

AdaMax \cite{kingma2014adam} generalizes Adam to the $\ell_\infty$-norm, stabilizing parameter updates. Replacing the squared gradients with the maximum absolute gradient values across time steps, AdaMax redefines the $v_t$ term using the infinity norm $u_t$:

\begin{equation}
    u_t = \max(\beta_2 \cdot u_{t-1}, |g_t|)
\end{equation}
The update rule for AdaMax then becomes:

\begin{equation}
    \theta_{t+1} = \theta_{t} - \frac{\eta}{u_t} \hat{m}_t
\end{equation}
This approach is beneficial when gradients vary greatly across parameters, as it provides a stable adaptation without requiring bias correction for $u_t$.

\subsection{Nadam}

Nadam \cite{dozat2016incorporating}, or Nesterov-accelerated Adam, incorporates Nesterov momentum into Adam’s update. In Nesterov’s method, the gradient is computed with respect to the look-ahead position. This configure enables better directional accuracy:
\begin{equation}
    \begin{aligned}
g_t &= \nabla_{\theta_t} J(\theta_t - \gamma m_{t-1}), \\
m_t &= \gamma m_{t-1} + \eta g_t,\\
\theta_{t+1} &= \theta_t - m_t,
\end{aligned}
\end{equation}
incorporating this look-ahead momentum, Nadam updates parameters using both the current and past gradient information:
\begin{equation}
    \theta_{t+1} = \theta_{t} - \frac{\eta}{\sqrt{\hat{v}_t} + \epsilon} \left( \beta_1 \hat{m}_t + \frac{(1 - \beta_1) g_t}{1 - \beta^t_1} \right)
\end{equation}
This modification allows Nadam to achieve faster convergence by accounting for momentum adjustments in advance. Each of these method has contributed significantly to optimizing neural network training by refining learning rates and stability. These updates allow making gradient descent more adaptable to complex datasets and models, in our work we proposed an extension of \textit{Adam} refining the learning rate of the algorithm using the characteristics of the networks.

\section{Algorithm}
We introduce three scaling methods: \textit{additive MinMaxMedian} scaling, \textit{multiplicative MinMaxMedian} scaling, and \textit{depth-based} scaling. Each strategy adjusts the effective learning factor for the neural network parameter according to its structural characteristics. This update allows faster and more stable training. Through a theoretical analysis, we show that these adaptive adjustments improve the speed of convergence compared to the classical Adam optimizer, by allowing a faster descent of the gradient and avoiding oscillations or too slow adjustments.

\subsection{Scaling}
The computation of scaling factor $S$ plays a crucial role in \textit{CaAdam}. For the \textit{additive MinMaxMedian} method, we define the scaling factor such

\begin{equation}
        S_{+}=\begin{cases}
     1 + \gamma   \frac{\Tilde{c}-c}{\Tilde{c}-c_{min}}, & \text{if $c\leq \Tilde{c} $ }.\\
     1 - \gamma   \frac{c - \Tilde{c}}{c_{max} - \Tilde{c}}, & \text{if $c > \Tilde{c} $}.
  \end{cases}
  \label{eq:scaling_additive}
\end{equation}

where $\gamma$ is an hyperparameter scaling factor, we set this value by default 0.95. $\Tilde{c}$ represents the median number of connections, which is used as a benchmark for whether the scaling should increase or decrease. When the current number of connections $c$ is less or equal to $\Tilde{c}$ the scaling factor using the $additive$ method is enhanced.The impact will be a more pronounced adjustment during the optimization process. The strength of this enhancement is control using $\gamma$ which can be customized. On the other hand if $c$ is greater than $\Tilde{c}$, $S_{+}$ will be lowered. The normalization of these adjustments is based on $c_{min}$ and $c_{max}$, which represent the minimum and maximum number of connections, respectively. These parameters help balance the optimization process by making the scaling adaptive, promoting a smoother and more balanced training across different layers of the network.
For the \textit{multiplicative MixMaxMedian} scaling method we first normalized the value of connection:

\begin{equation}
        \sigma =\begin{cases}
     \frac{\Tilde{c}-c}{\Tilde{c}-c_{min}}, & \text{if $c\leq \Tilde{c} $ }.\\
     \frac{c - \Tilde{c}}{c_{max} - \Tilde{c}}, & \text{if $c > \Tilde{c} $}.
  \end{cases}
\end{equation}

Once obtained, we will update $S$, which will be defined as:

\begin{equation}
    S_{*} = \exp^{ \sigma \log (\gamma)}
\end{equation}

where $\gamma$ denotes the same scaling factor introduced Eq.\eqref{eq:scaling_additive}

\begin{equation}
    S_{d} = (1+\gamma)^{\frac{d_m - (1+d)}{d_m}},
\end{equation}
where $d_m$ denotes the total depth of the neural network, $d$ the depth of the current layer. The equation adjusts  $S_a$ based on the depth of the layer. Layers that are closer to the input receive a scaling factor that is more influenced by $\gamma$, while deeper layers have a diminished scaling effect. This helps manage how gradients are adjusted across different depths, ensuring that adjustments are not uniformly applied across the network, thereby encouraging a more balanced training process across all layers of the model.

\subsection{Adam}

Adaptive Moment Estimation (Adam) \cite{kingma2014adam}  computes adaptive learning rates for each parameter. In addition to storing an exponentially decaying average of past squared gradients $v_t$ Adam keeps an exponentially decaying average of past gradients $m_t$, similar to momentum:
\begin{equation}
    \begin{split}
m_t &= \beta_1 m_{t-1} + (1 - \beta_1) g_t,\\
v_t &= \beta_2 v_{t-1} + (1 - \beta_2) g_t^2,
\end{split}
\end{equation}
where $m_t$ and $v_t$ are estimates of the first moment (the mean) and the second moment (the uncentered variance) of the gradients respectively, hence the name of the method. As $m_t$ and $v_t$ are initialized as vectors of $0$'s, the authors of Adam observe that they are biased towards zero, especially during the initial time steps, and especially when the decay rates are small (i.e. $\beta_1$ and $\beta_2$ are close to $1$). The authors \cite{kingma2014adam} recommend using default values of $\beta_1 = 0.9$, $\beta_2 = 0.999$, and $\epsilon = 10^{-8}$. Through empirical evaluation, they demonstrate that Adam performs effectively in practice and offers competitive results compared to other adaptive learning-rate algorithms. They counteract these biases by computing bias-corrected first and second moment estimates:

\begin{equation}
\begin{split}
\hat{m}_t &= \frac{m_t}{1 - \beta^t_1},\\
\hat{v}_t &= \frac{v_t}{1 - \beta^t_2}.
\end{split}
\end{equation}

\subsection{Update}
They then use these to update the parameters just as we have seen in Adadelta and RMSprop, which yields the Adam update rule:
\begin{equation}
\theta_{t+1} = \theta_{t} - \alpha \frac{\hat{m}_t}{\sqrt{\hat{v}_t} + \epsilon} 
\label{eq:adam_update}
\end{equation}
Our method introduce a \textit{scaling factor} denoted $S$ which modify the learning rate update, the general update step \eqref{eq:adam_update} will become:
\begin{equation}
\theta_{t+1} = \theta_{t} - \alpha \cdot S \cdot \frac{\hat{m}_t}{\sqrt{\hat{v}_t} + \epsilon} 
\label{eq:adam_update}
\end{equation}
where $S$ is calculated based on specific scaling strategies; additive scaling (linearly on connection counts or depth) and multiplicative scaling (uses exponential functions based on the connection structure). Looking at the impact on the learning rate $a$, the effective learning rate using \textit{CaAdam} $\Tilde{\alpha} = \alpha \cdot S$. The critical insight is that $S$ dynamically adjusts $\Tilde{\alpha}$ based on structural properties of the networks (layer depth, connection counts), which can make learning more efficient. 

\subsection{A view on convergence effect}
Now, we need to make sure that the expected value of the squared gradient norm $\mathbb{E}\left[\|\nabla f(\theta)\|^2\right]$ will decrease faster or at least at the same speed. The decline in the loss function can be described as
\begin{equation}
    f(\theta_{t+1}) - f(\theta_t) \approx \nabla f(\theta_t) \cdot (\theta_{t+1} - \theta_t),
    \label{eq:loss}
\end{equation}
replacing $(\theta_{t+1} - \theta_t)$ from Eq.\eqref{eq:loss} using \textit{Adam} update Eq.\eqref{eq:adam_update} we have
\begin{equation}
    f(\theta_{t+1}) - f(\theta_t) \approx -\alpha \cdot \frac{\nabla f(\theta_t) \cdot \hat{{m}}_t}{\sqrt{\hat{{v}}_t} + \epsilon}.
\end{equation}
Taking the expection we obtain:
\begin{equation}
    \mathbb{E}[f(\theta_{t+1})] \leq \mathbb{E}[f(\theta_t)] - \alpha \cdot \mathbb{E} \left[\frac{\|\nabla f(\theta_t)\|^2}{\sqrt{\hat{{v}}_t} + \epsilon}\right].
    \label{eq:expectation_adam}
\end{equation}
Using our scaling factor $S$, we would have 
\begin{equation}
    f(\theta_{t+1}) - f(\theta_t) \approx -\alpha \cdot S \cdot \frac{\nabla f(\theta_t) \cdot \hat{{m}}_t}{\sqrt{\hat{{v}}_t} + \epsilon},
\end{equation}
and taking the expectation over the distribution of gradients we obtain
\begin{equation}
    \mathbb{E}[f(\theta_{t+1})] \leq \mathbb{E}[f(\theta_t)] - \alpha \cdot \mathbb{E} \left[S \cdot \frac{\|\nabla f(\theta_t)\|^2}{\sqrt{\hat{{v}}_t} + \epsilon}\right].
    \label{eq:expectation_caadam}
\end{equation}
For parts of the model where $S>1$ (e.g. layers needing quicker adaptation or sparse gradients), the effective learning rate is higher $\Tilde{\alpha}>\alpha$ and leads to bigger steps to reach the optimal solution. This case will allow to reduce the loss faster, on the other side if $S<1$ the step size will be reduced $\Tilde{\alpha}<\alpha$ and help the optimizer to avoid oscillations why may occur when gradient are noisy. Adapting the learning rate locally will allow to converge without large deviations. Looking at Eq.\eqref{eq:expectation_adam}-\eqref{eq:expectation_caadam} when we compare the convergence rate, $S$ adjusts adaptively based on structural information of the networks. This extension ensures $\mathbb{E}\left[S\right]\geq 1$ when acceleration is needed leading to:
\begin{equation}
    \mathbb{E}[f(\theta_{t+1})] - \mathbb{E}[f(\theta_t)]\leq - \alpha \cdot \Bar{s} \cdot \mathbb{E} \left[ \frac{\|\nabla f(\theta_t)\|^2}{\sqrt{\hat{{v}}_t} + \epsilon}\right],
\end{equation}
where $\Bar{s} \geq 1$ on average leading to faster descent. Custom scaling strategies embedded in the optimizer lead to faster convergence under typical conditions, as long as the scaling factor $S$ is designed properly to respond adaptively to the neural networks architecture and training dynamics.
\section{Experimental Setup}

\subsection{Learning Tasks and Datasets}
To rigorously validate the effectiveness of CaAdam, we constructed a comprehensive evaluation framework spanning both classification and regression domains. Our experimental design incorporated three classification tasks and one regression task.

\medskip

For classification, we employed the CIFAR-10, CIFAR-100 \cite{krizhevsky2009learning}, and Fashion-MNIST datasets \cite{xiao2017fashion}. The CIFAR datasets comprise 60,000 natural color images at 32×32 resolution, with 50,000 allocated for training and 10,000 for testing. While CIFAR-10 categorizes images into 10 classes, CIFAR-100 presents a more challenging scenario with 100 classes, allowing us to evaluate our optimizer's performance on tasks of varying complexity. For the Fashion-MNIST dataset, we performed preprocessing steps by upscaling the original 28×28 images to 56×56 and converting to RGB format using bicubic interpolation, maintaining the original 10-class categorization of fashion items.

\medskip
For regression analysis, we selected the California Housing dataset, which contains 20,640 samples describing housing districts in California. Each sample comprises 8 numerical features including median income, housing median age, and geographical coordinates, with the median house value as the target variable.
We relied on Keras dataset directly as our data source.

\medskip

\subsection{Model Architectures}

\medskip

Our architectural choices were driven by the need for consistency in classification tasks while allowing varied complexity in regression scenarios. For all image classification tasks (CIFAR-10, CIFAR-100, and Fashion-MNIST), we implemented a ResNet-20 architecture, as using pre-trained larger models would not have been appropriate, and trying to calibrate such big models from scratch on relatively small datasets would be impractical. This network follows the standard residual block pattern with skip connections, facilitating better gradient flow during training. The architecture processes 32×32×3 input images for CIFAR datasets and 56×56×3 for the upscaled Fashion-MNIST data. The only architectural variation across these classification tasks lies in the output layer, which contains 10 units for CIFAR-10 and Fashion-MNIST, and 100 units for CIFAR-100, corresponding to their respective number of classes.

\medskip

For the regression task, we employed multiple Multi-Layer Perceptron (MLP) configurations of varying complexity. We systematically explored architectures ranging from shallow networks (two layers: 64→32 units) to deeper configurations (four layers: 1024→256→64→16 units). This systematic variation in network depth and width enables us to analyze how our connection-aware scaling strategies perform across different network topologies.

\subsection{Training Protocol and Error Metrics}
Our training protocol was designed to ensure robust optimization while preventing overfitting. We employed the standard Mean Squared Error (MSE) loss function for regression tasks and sparse categorical cross-entropy loss for classification tasks, as these are the de facto objectives for their respective tasks in the deep learning literature. This choice allows for a fair evaluation of our optimizer against existing methods on these canonical loss landscapes.
The implementation of early stopping with a patience of 15 epochs serves two crucial purposes: it prevents overfitting by monitoring validation loss with a minimum delta of 1e-5, and it ensures fair comparison of convergence speeds across different optimizers. The choice of this particular patience value balances the trade-off between allowing sufficient time for convergence and preventing unnecessary computation.

\medskip

Our learning rate scheduler employs a reduction factor of 0.25 with a patience of 6 epochs, triggering when validation loss plateaus. This aggressive reduction factor, combined with a minimum learning rate of 2.5e-5, allows the optimizer to make rapid progress initially while ensuring fine-grained optimization in later stages. The minimum learning rate was carefully chosen to prevent premature convergence while maintaining numerical stability.
For evaluation metrics, we utilize Root Mean Square Error (RMSE) for regression tasks, calculated as:

\begin{equation}
RMSE = \sqrt{\frac{1}{n}\sum_{i=1}^{n}(y_i - \hat{y}_i)^2}
\end{equation}

\medskip

As with our loss functions, we selected these standard metrics to facilitate direct comparisons with existing optimization methods: RMSE for regression tasks and accuracy for classification tasks. We also track training time and epochs to convergence to assess computational efficiency, as these are crucial performance indicators for optimization algorithms.

\subsection{Comparative Analysis Framework}

To establish the effectiveness of CaAdam, we conducted a comprehensive comparison against four state-of-the-art optimizers: standard Adam, AdamW (Adam with decoupled weight decay), Adamax (utilizing the infinity norm), and Nadam (incorporating Nesterov momentum). Each experiment was repeated 30 times to ensure statistical robustness, with all optimizers initialized with a learning rate of 0.001 and a batch size of 64. The maximum number of epochs was set to 1,000, though early stopping typically triggered well before this limit. Our CaAdam implementation was evaluated with three distinct scaling strategies: additive MinMaxMedian scaling, multiplicative MinMaxMedian scaling, and depth-based scaling as described in section III. Statistical significance was assessed through independent t-tests comparing each optimizer variant against the Adam baseline. We employed three significance thresholds (p < 0.05, p < 0.01, and p < 0.001) to provide a nuanced view of performance differences. Training time and convergence speed were analyzed alongside accuracy/RMSE to provide a comprehensive evaluation of each optimizer's practical utility. The consistent use of ResNet-20 across all classification tasks ensures that performance differences can be attributed to the optimizers' behavior rather than architectural variations, while the diverse MLP configurations for regression allow us to evaluate our optimizer's adaptability to different network structures.

\section{Results and Analysis}

\subsection{Overview}
We present our experimental results through a comprehensive analysis of both classification and regression tasks. Tables~\ref{table:california_raw_results}, \ref{table:classification_raw_results} provide the raw performance metrics, while Tables~\ref{table:california_statistical_results}, \ref{table:classification_statistical_results} present the statistical analysis of improvements over the Adam baseline.

\subsection{Classification Performance}
On image classification tasks, CaAdam demonstrated consistent improvements over the baseline Adam optimizer across all datasets. Most notably, on CIFAR-10, the multiplicative scaling strategy achieved an accuracy of 83.1\% ($\pm$0.5\%), representing a statistically significant improvement of 4.09\% ($p < 0.001$) over Adam's 79.8\% baseline. This improvement was achieved while maintaining comparable training times, suggesting that the enhanced performance does not come at the cost of computational efficiency.

\medskip

For the more challenging CIFAR-100 dataset, both depth-based and multiplicative scaling strategies showed substantial improvements, achieving accuracies of 50.4\% and 50.7\% respectively, compared to Adam's 47.8\%. The multiplicative strategy's 5.97\% improvement was particularly noteworthy ($p < 0.001$), demonstrating CaAdam's effectiveness in handling complex classification tasks with numerous classes.

\medskip

On Fashion-MNIST, while the absolute improvements were more modest due to the already high baseline performance, CaAdam with depth-based scaling still achieved a statistically significant improvement of 0.39\% ($p < 0.001$), reaching 92.6\% accuracy compared to Adam's 92.3\%. This suggests that our connection-aware approach can provide benefits even in scenarios where conventional optimizers already perform well.

\subsection{Regression Analysis}
The regression results on the California Housing dataset reveal interesting patterns across different MLP architectures. Most significantly, CaAdam with multiplicative scaling consistently outperformed other optimizers across all architectural configurations, with improvements in RMSE ranging from 1.43\% to 2.87\% (all statistically significant at $p < 0.001$).

\medskip

The most substantial improvement was observed in the [128, 64, 32] architecture, where multiplicative scaling achieved an RMSE of 0.446 ($\pm$0.003) compared to Adam's 0.459 ($\pm$0.003), representing a 2.87\% improvement. Notably, this enhancement was accompanied by a 15.41\% reduction in training time ($p < 0.01$), demonstrating that CaAdam can simultaneously improve both accuracy and efficiency.

\medskip

An interesting pattern emerged in the relationship between architectural complexity and optimizer performance. For simpler architectures ([64, 32]), CaAdam showed particularly strong improvements in both RMSE (2.07\% improvement, $p < 0.001$) and training time (30.11\% reduction, $p < 0.01$). This suggests that our connection-aware approach is especially effective at optimizing smaller networks, where each parameter's contribution is more significant.

\subsection{Convergence Characteristics}
Across both classification and regression tasks, CaAdam typically required fewer epochs to converge compared to the baseline Adam optimizer. This is particularly evident in the regression tasks, where the multiplicative scaling strategy reduced the number of epochs to convergence by up to 34.6\% for certain architectures. This faster convergence did not come at the cost of final model performance, suggesting that CaAdam's scaling strategies help navigate the loss landscape more efficiently.

\medskip

\subsection{Scaling Strategy Comparison}
Among the three proposed scaling strategies, multiplicative scaling demonstrated the most consistent performance improvements across all tasks. While additive and depth-based scaling also showed significant improvements over the baseline, multiplicative scaling's superior performance can be attributed to its more nuanced handling of parameter updates based on layer connectivity patterns. 

\medskip

The depth-based strategy showed particular strength in deeper architectures, especially in the CIFAR-100 classification task, suggesting its utility in scenarios with more complex network hierarchies.

\medskip

This performance pattern aligns with the foundational design principles of each scaling strategy. The multiplicative and additive scaling approaches were initially conceived for simpler MLP architectures, where they can leverage straightforward proxies like connection counts to capture layer-wise optimization needs. These strategies excel in such contexts because the relationship between layer connectivity and optimal learning rates tends to be more direct and predictable in simpler networks. 

\medskip

In contrast, the depth-based scaling strategy was designed specifically for deeper architectures, using layer depth as a simple yet effective proxy for optimization behavior. While this is a more basic architectural indicator compared to connection counting, it proves particularly effective in deeper networks where vanishing gradient issues are more prominent. This explains its strong performance on CIFAR-100 with ResNet20, where the depth-based approach can help counteract the challenges of propagating gradients through many layers.

\medskip

These complementary approaches demonstrate how even relatively simple architectural proxies can significantly improve optimization when properly matched to network characteristics. Our results suggest that future research could explore more sophisticated architectural indicators, potentially combining multiple structural features to create even more effective scaling strategies.
\subsection{Training Loss Convergence Analysis}
In order to better visualize optimizer effectiveness, we analyzed training loss evolution across different network architectures. Each configuration was tested with 5 independent runs to ensure representativeness while maintaining clarity. Individual runs were plotted separately to preserve information about specific optimization paths.
\subsubsection{Small MLP Architectures}
\begin{figure}[H]
    \centering
    \includegraphics[width=0.4\textwidth]{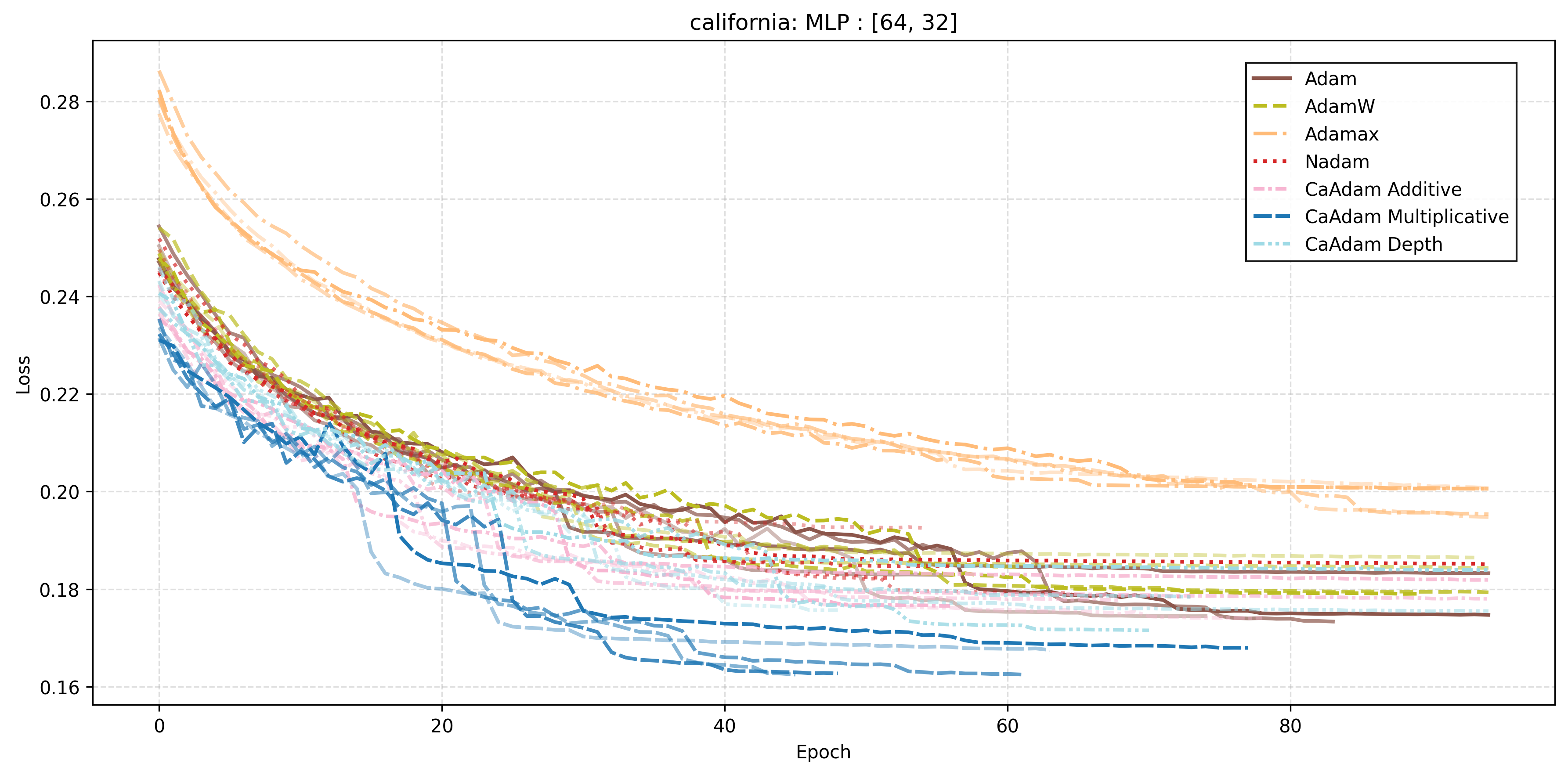}
    \caption{Training loss convergence on California Housing dataset with MLP [64, 32]}
    \label{fig:california_mlp_small}
\end{figure}
\begin{figure}[H]
    \centering
    \includegraphics[width=0.4\textwidth]{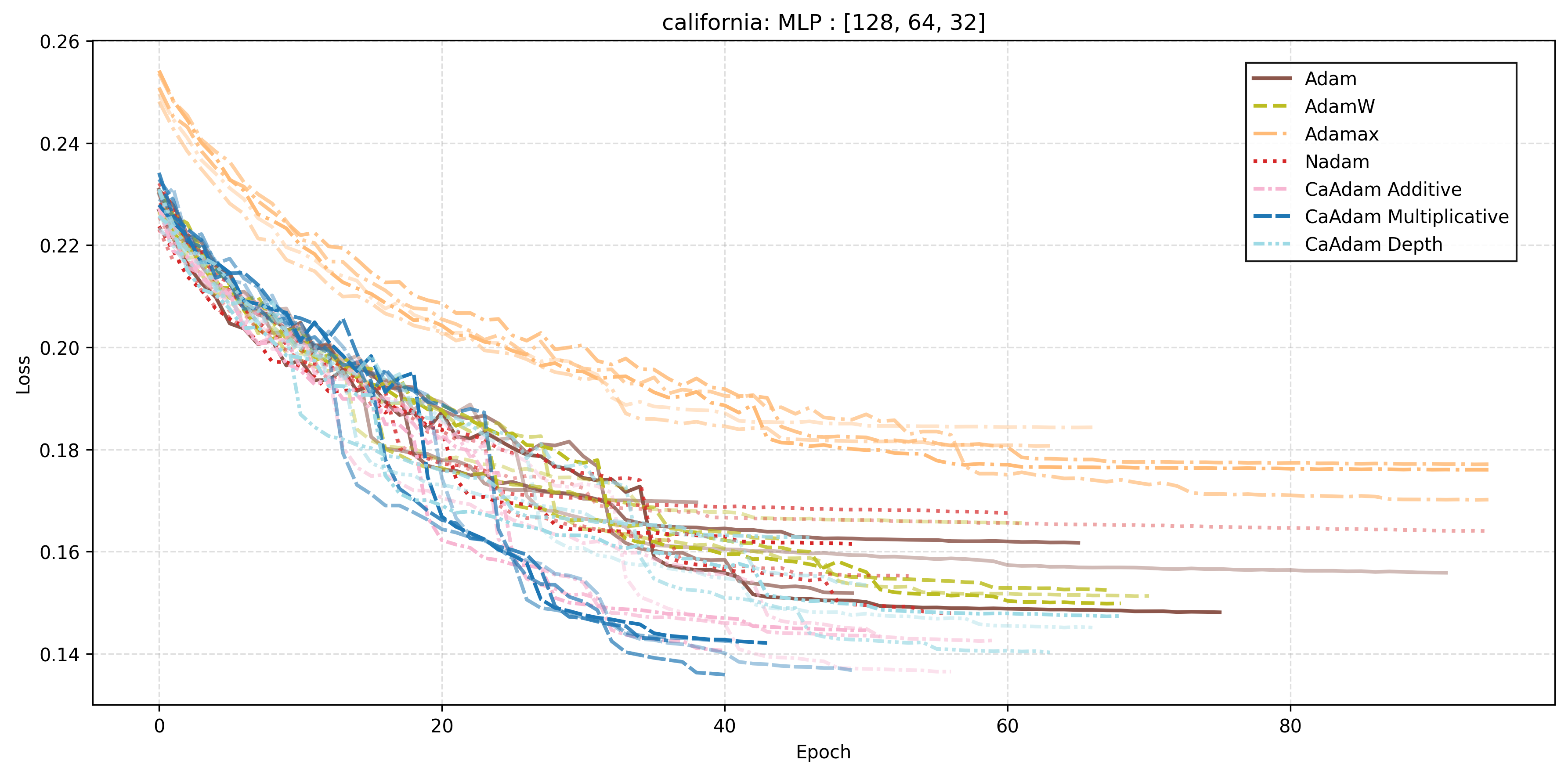}
    \caption{Training loss convergence on California Housing dataset with MLP [128, 64, 32]}
    \label{fig:california_mlp_medium}
\end{figure}
\begin{figure}[H]
    \centering
    \includegraphics[width=0.4\textwidth]{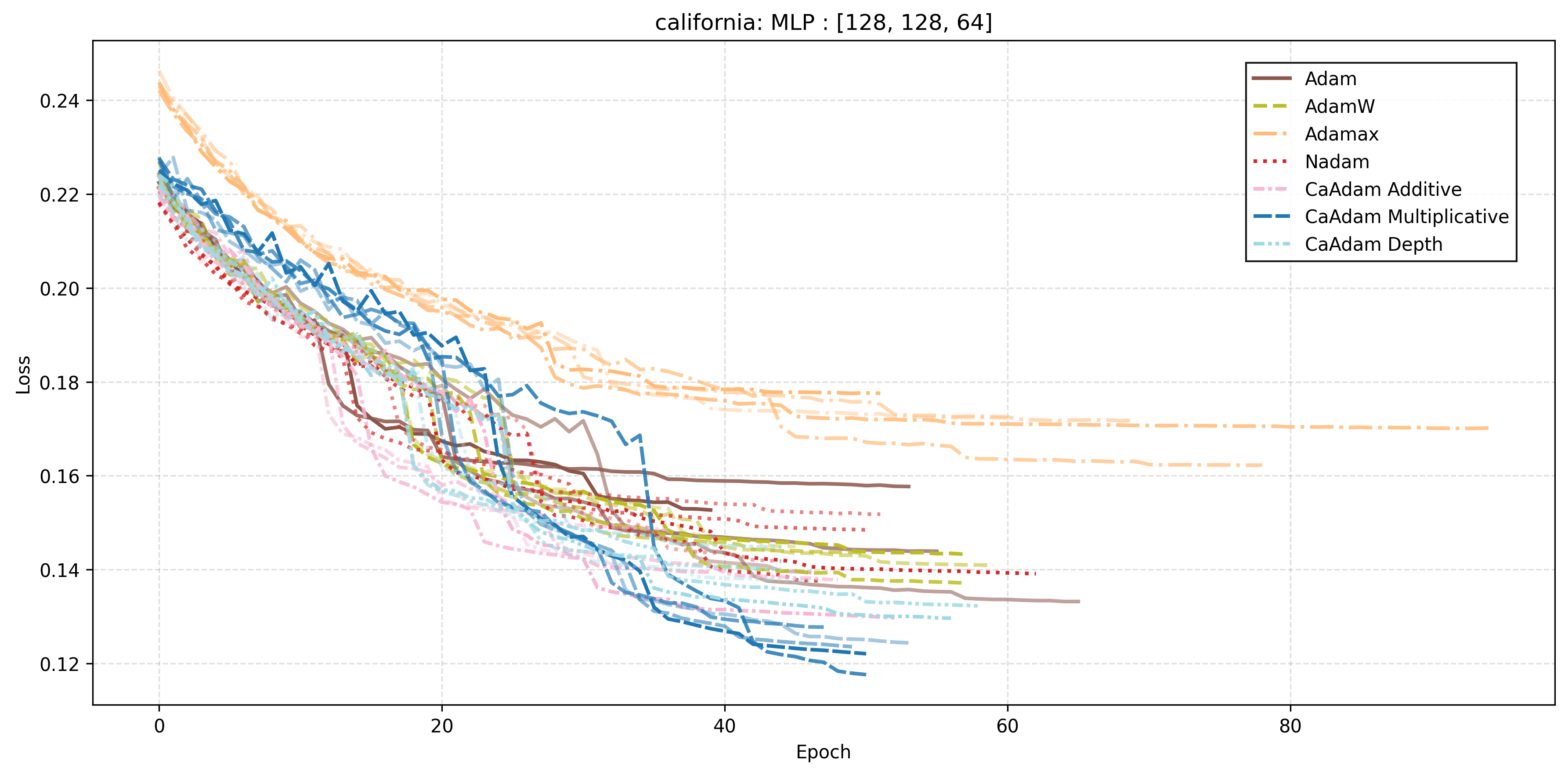}
    \caption{Training loss convergence on California Housing dataset with MLP [128, 128, 64]}
    \label{fig:california_mlp_wide}
\end{figure}
For the smaller MLP configurations ([64, 32], [128, 64, 32], and [128, 128, 64]) (Figures \ref{fig:california_mlp_small}--\ref{fig:california_mlp_wide}), CaAdam with multiplicative scaling consistently demonstrated superior performance, achieving both faster convergence and lower final loss values. This aligns with our design intentions, as multiplicative scaling was optimized for simpler architectures where layer relationships can be effectively captured through basic weight statistics. The advantage of multiplicative scaling became particularly evident around epoch 20 across these architectures, suggesting enhanced effectiveness during fine-tuning phases. Interestingly, as we moved to wider architectures like [128, 128, 64], both additive and depth-based scaling showed competitive early-stage performance, indicating that increased network width affects the relative strengths of different scaling strategies during different training phases.
\subsubsection{Large MLP Architectures}
\begin{figure}[H]
    \centering
    \includegraphics[width=0.4\textwidth]{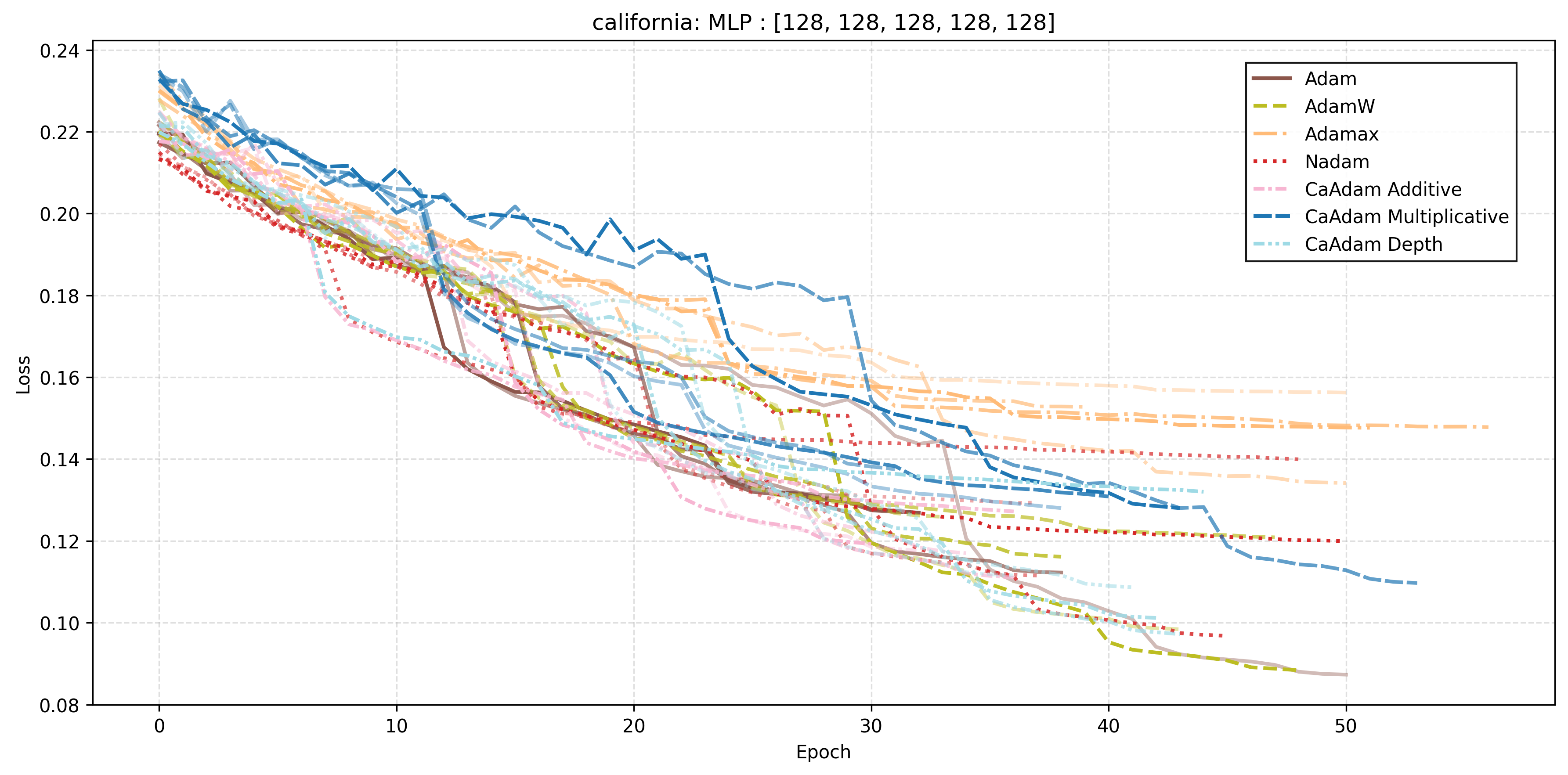}
    \caption{Training loss convergence on California Housing dataset with MLP [128, 128, 128, 128, 128]}
    \label{fig:california_mlp_deep}
\end{figure}
\begin{figure}[H]
    \centering
    \includegraphics[width=0.4\textwidth]{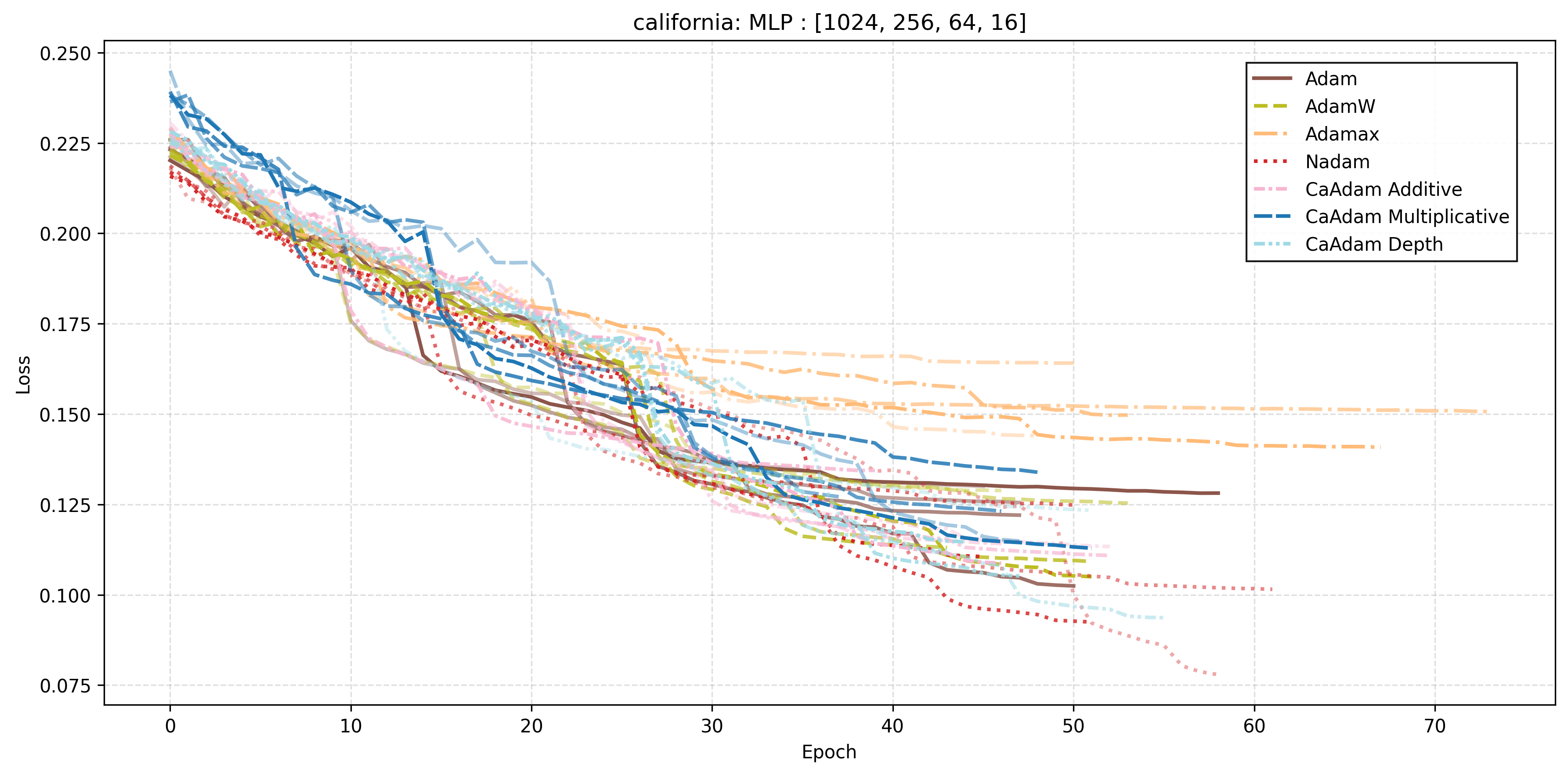}
    \caption{Training loss convergence on California Housing dataset with MLP [1024, 256, 64, 16]}
    \label{fig:california_mlp_large}
\end{figure}
The behavior patterns evolved significantly in larger architectures (Figures \ref{fig:california_mlp_deep}, \ref{fig:california_mlp_large}). In these deeper configurations, convergence paths showed increased oscillation. Although multiplicative scaling maintained its overall effectiveness, depth-based scaling demonstrated notably improved relative performance, aligning with its design goals for deeper networks.

\medskip

In the most complex architecture [1024, 256, 64, 16], the optimizer dynamics became more nuanced. Multiplicative scaling exhibited slower initial progress but maintained steady improvement throughout training. The strong performance of depth-based scaling in this context validates its design principles for complex network hierarchies, suggesting that our different scaling strategies complement each other across varying network complexities. These results demonstrate how our different scaling approaches adapt to network depth and width, with multiplicative scaling excelling in simpler architectures while depth-based approaches show increasing benefits as network complexity grows. The consistent performance improvements across all architectures validate our connection-aware approach to optimization.

\subsubsection{ResNet20 on CIFAR Datasets}
The ResNet20 experiments on CIFAR datasets revealed distinct patterns:
\begin{figure}[H]
    \centering
    \includegraphics[width=0.4\textwidth]{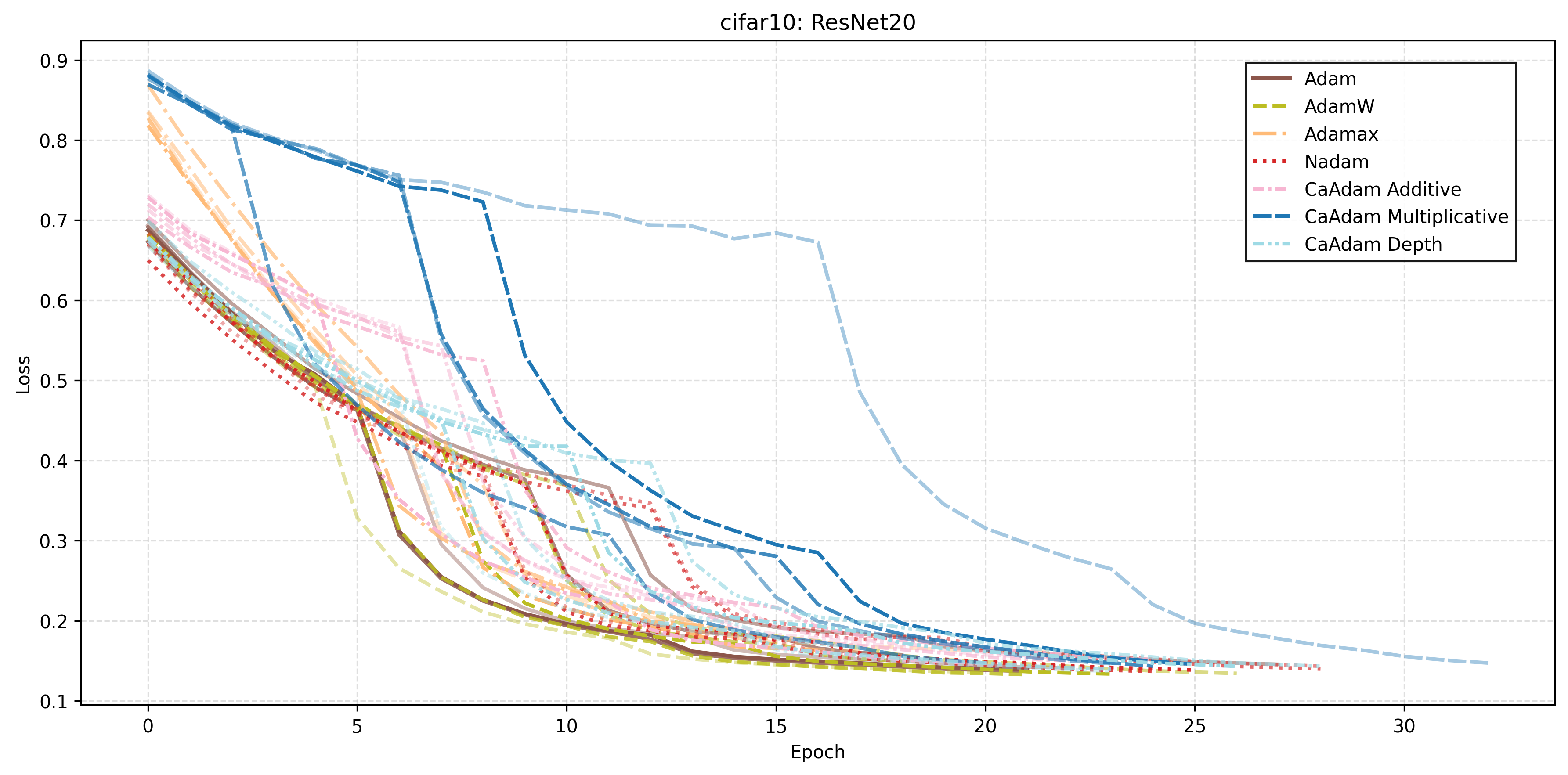}
    \caption{Training loss convergence on CIFAR-10 with ResNet20}
    \label{fig:cifar10_convergence}
\end{figure}

\begin{figure}[H]
    \centering
    \includegraphics[width=0.4\textwidth]{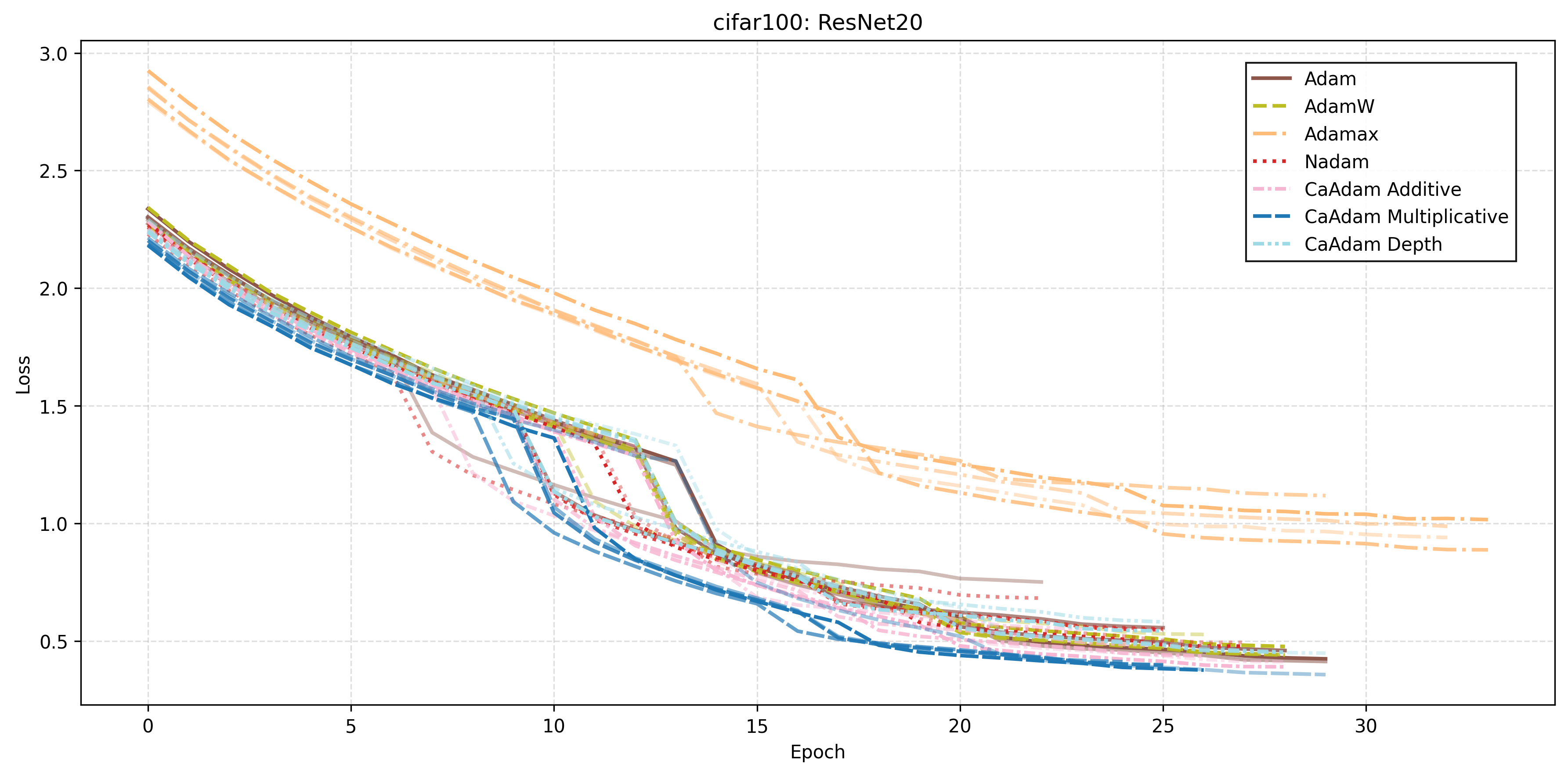}
    \caption{Training loss convergence on CIFAR-100 with ResNet20}
    \label{fig:cifar100_convergence}
\end{figure}
On CIFAR-10 (Figure \ref{fig:cifar10_convergence}), both multiplicative and depth-based scaling initially showed higher loss values but demonstrated steeper descent rates after epoch 5. This suggests more effective loss landscape navigation once the optimization process stabilizes, despite these approaches being originally designed for simpler architectures.

\medskip

CIFAR-100's more challenging nature (Figure \ref{fig:cifar100_convergence}) highlighted the strength of depth-based scaling, which achieved notably lower final loss values. This superior performance aligns with the strategy's design focus on deeper architectures.
\subsubsection{Fashion-MNIST Results}
\begin{figure}[H]
    \centering
    \includegraphics[width=0.4\textwidth]{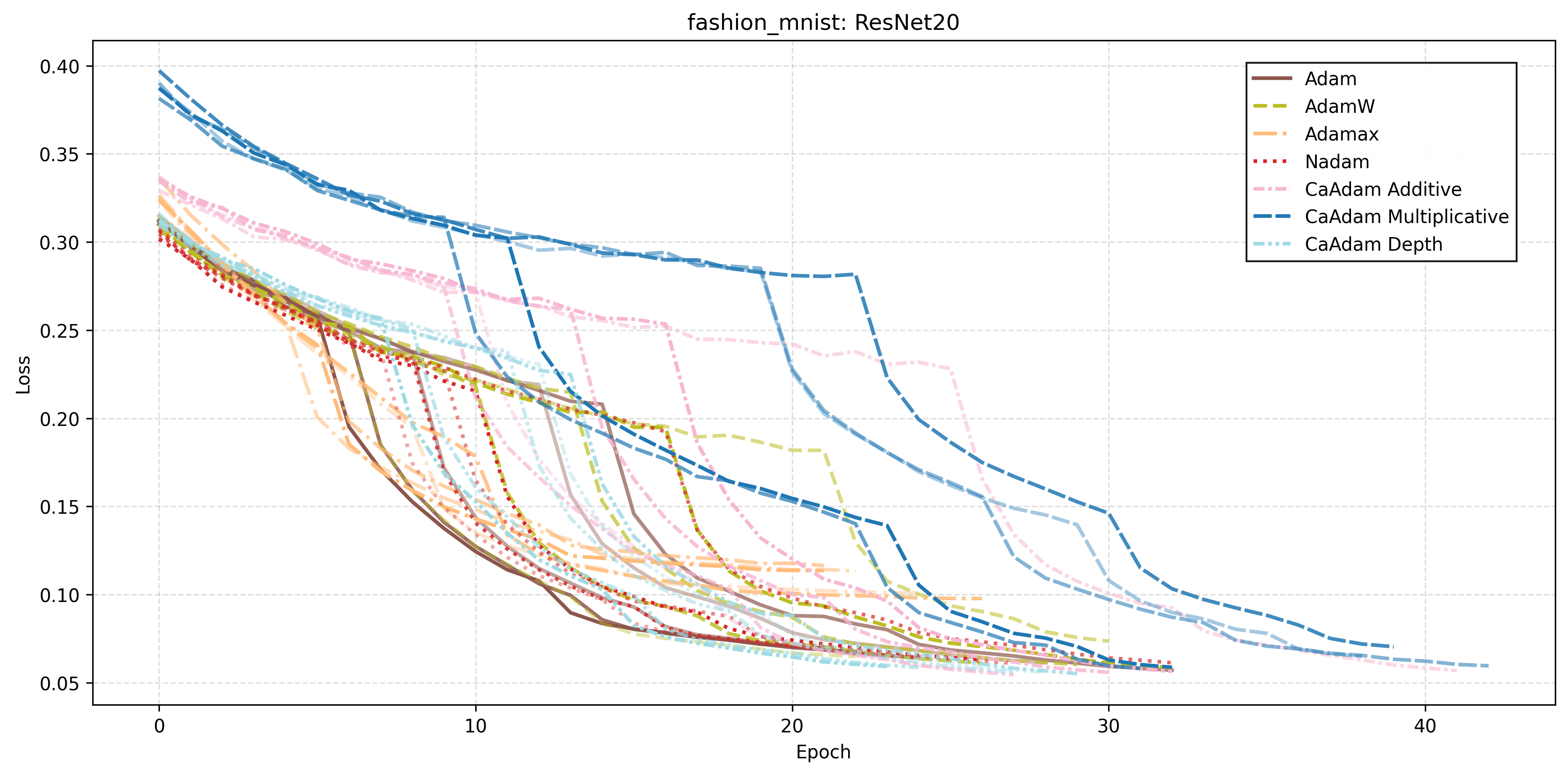}
    \caption{Training loss convergence on Fashion-MNIST with ResNet20}
    \label{fig:fashion_mnist_convergence}
\end{figure}
The Fashion-MNIST experiments with ResNet20 (Figure \ref{fig:fashion_mnist_convergence}) revealed a distinctive pattern where all CaAdam variants initially maintained higher loss values but achieved superior final convergence. This behavior suggests that our connection-aware scaling strategies excel at fine-tuning network parameters in later training stages, even when applied to ResNet architectures.

\medskip

This comprehensive analysis demonstrates how different CaAdam variants adapt to varying network architectures and task complexities, with multiplicative scaling generally excelling in simpler architectures and depth-based approaches showing increasing benefits as network complexity grows.

\section{Conclusion}
This work introduces CaAdam, a novel optimization approach that challenges the traditional paradigm of architecture-agnostic parameter updates in neural networks. By incorporating structural information through simple scaling strategies, we have demonstrated that connection-aware optimization can significantly improve training dynamics across diverse architectures and tasks. Our empirical results consistently show the benefits of this approach across both classification and regression tasks. The improvements in accuracy and reduction in training time demonstrate that incorporating architectural awareness into optimization strategies can enhance both the efficiency and effectiveness of neural network training. Moreover, these improvements were achieved while working within the constraints of current deep learning frameworks, suggesting the practical applicability of our method. The complementary nature of our scaling strategies emerged as a key finding. Multiplicative scaling proved particularly effective for simpler architectures, where connection patterns directly inform optimal learning rates. In contrast, depth-based scaling showed its strength in deeper networks, especially evident in its superior performance on ResNet architectures. This pattern suggests that different architectural characteristics may require distinct optimization approaches, challenging the one-size-fits-all philosophy of traditional optimizers. While our current implementation relies on relatively simple architectural proxies, the consistent improvements across various tasks and architectures suggest a promising direction for future research. More sophisticated approaches might consider dynamic adaptation of scaling strategies during training, integration of layer-type specific optimization behaviors, or incorporation of more complex topological features. The potential also exists for developing systems that automatically select and adjust scaling strategies based on comprehensive architecture analysis. In addition, preliminary testing has shown more mixed results when using this method for Recurrent Neural Networks (RNNs), and we would not advise using our work as-is for this kind of network, though we encourage more research to adapt the idea to them.

\medskip

\textit{We believe this work opens new avenues for research in architecture-aware optimization, suggesting that the future of neural network training lies in methods that more closely align with the structural characteristics of their target networks. The success of our simple proxies indicates that even basic architectural awareness can significantly impact optimization performance, pointing toward potentially greater improvements as more sophisticated approaches are developed.}

\bibliographystyle{IEEEtran}
\bibliography{bib}

\newpage
\appendices
\onecolumn
\begin{table*}[!h]
\centering
\caption{Regression Results on California Housing Dataset}
\label{tab:regression_results}
\resizebox{0.8\textwidth}{!}{%
\begin{tabular}{lllcccccccc}
\toprule
\textbf{Architecture} & \textbf{Optimizer} & \textbf{Scaling Strategy} & \textbf{RMSE} & \textbf{RMSE Std} & \textbf{Time Mean} & \textbf{Time Std} & \textbf{Epochs Mean} & \textbf{Epochs Std} \\
\midrule
\multirow{7}{*}{[1024, 256, 64, 16]} & Adam & - & 0.442 & 0.002 & 14.84 & 1.73 & 72.50 & 8.81 \\
& AdamW & - & 0.442 & 0.003 & 14.82 & 1.85 & 72.83 & 9.65 \\
& Adamax & - & 0.452 & 0.002 & 18.74 & 3.55 & 93.00 & 18.19 \\
& CAdam & Additive & \textbf{0.440} & 0.003 & 14.23 & 1.44 & 69.00 & 7.03 \\
& CAdam & Depth & \underline{\textbf{0.440}} & 0.003 & 14.49 & 1.69 & 71.03 & 9.05 \\
& CAdam & Multiplicative & 0.443 & 0.004 & 14.07 & 1.12 & 68.87 & 5.82 \\
& Nadam & - & 0.441 & 0.003 & 15.85 & 2.83 & 75.77 & 14.65 \\
\midrule
\multirow{7}{*}{[128, 128, 128, 128, 128]} & Adam & - & 0.455 & 0.003 & 6.85 & 0.63 & 61.33 & 6.13 \\
& AdamW & - & 0.453 & 0.003 & 6.75 & 0.55 & 58.27 & 5.46 \\
& Adamax & - & 0.464 & 0.002 & 8.14 & 0.82 & 72.60 & 8.30 \\
& CAdam & Additive & 0.450 & 0.004 & 6.60 & 0.60 & 56.83 & 5.66 \\
& CAdam & Depth & 0.450 & 0.003 & 6.83 & 0.57 & 59.87 & 5.66 \\
& CAdam & Multiplicative & \textbf{0.448} & 0.003 & 7.05 & 0.67 & 61.13 & 7.24 \\
& Nadam & - & 0.454 & 0.003 & 7.27 & 0.76 & 60.80 & 7.71 \\
\midrule
\multirow{7}{*}{[128, 128, 64]} & Adam & - & 0.456 & 0.002 & 4.86 & 0.45 & 76.57 & 7.44 \\
& AdamW & - & 0.456 & 0.001 & 4.83 & 0.45 & 77.07 & 8.28 \\
& Adamax & - & 0.464 & 0.002 & 6.56 & 0.99 & 111.53 & 19.47 \\
& CAdam & Additive & 0.451 & 0.003 & 4.45 & 0.40 & 68.83 & 6.89 \\
& CAdam & Depth & 0.452 & 0.002 & 4.67 & 0.40 & 73.33 & 6.47 \\
& CAdam & Multiplicative & \textbf{0.446} & 0.003 & 4.49 & 0.29 & 69.50 & 5.30 \\
& Nadam & - & 0.456 & 0.002 & 4.92 & 0.51 & 75.33 & 9.13 \\
\midrule
\multirow{7}{*}{[128, 64, 32]} & Adam & - & 0.459 & 0.003 & 3.93 & 1.04 & 88.27 & 28.71 \\
& AdamW & - & 0.459 & 0.003 & 3.90 & 0.52 & 86.50 & 14.73 \\
& Adamax & - & 0.467 & 0.002 & 6.34 & 3.86 & 155.07 & 106.09 \\
& CAdam & Additive & 0.451 & 0.003 & 3.55 & 0.49 & 77.37 & 13.81 \\
& CAdam & Depth & 0.454 & 0.002 & 3.92 & 0.65 & 86.30 & 16.95 \\
& CAdam & Multiplicative & \textbf{0.446} & 0.003 & 3.33 & 0.28 & 70.73 & 8.66 \\
& Nadam & - & 0.460 & 0.002 & 3.87 & 0.57 & 80.87 & 15.06 \\
\midrule
\multirow{7}{*}{[256, 128, 64, 32]} & Adam & - & 0.452 & 0.002 & 5.75 & 0.73 & 68.83 & 10.64 \\
& AdamW & - & 0.451 & 0.003 & 5.81 & 0.35 & 68.57 & 4.92 \\
& Adamax & - & 0.459 & 0.002 & 7.62 & 1.37 & 96.47 & 19.70 \\
& CAdam & Additive & 0.447 & 0.004 & 5.46 & 0.53 & 64.47 & 6.95 \\
& CAdam & Depth & 0.448 & 0.003 & 5.73 & 0.50 & 68.40 & 7.04 \\
& CAdam & Multiplicative & \textbf{0.446} & 0.004 & 5.76 & 0.64 & 67.93 & 8.47 \\
& Nadam & - & 0.454 & 0.002 & 5.94 & 0.55 & 67.43 & 7.93 \\
\midrule
\multirow{7}{*}{[64, 32]} & Adam & - & 0.466 & 0.003 & 5.16 & 2.16 & 170.43 & 81.35 \\
& AdamW & - & 0.466 & 0.003 & 4.57 & 2.04 & 147.23 & 75.98 \\
& Adamax & - & 0.472 & 0.002 & 9.24 & 5.66 & 322.87 & 207.91 \\
& CAdam & Additive & 0.463 & 0.003 & 3.82 & 1.12 & 119.00 & 40.12 \\
& CAdam & Depth & 0.464 & 0.004 & 4.29 & 1.80 & 138.30 & 66.84 \\
& CAdam & Multiplicative & \textbf{0.446} & 0.003 & 3.61 & 1.66 & 111.87 & 61.85 \\
& Nadam & - & 0.465 & 0.003 & 4.22 & 1.37 & 130.17 & 50.47 \\
\bottomrule
\end{tabular}
}
\label{table:california_raw_results}
\end{table*}

\begin{table*}[!h]
\centering
\caption{Classification Results on CIFAR and Fashion-MNIST Datasets}
\label{tab:classification_results}
\resizebox{0.8\textwidth}{!}{%
\begin{tabular}{lllcccccccc}
\toprule
\textbf{Dataset} & \textbf{Architecture} & \textbf{Optimizer} & \textbf{Scaling Strategy} & \textbf{Accuracy} & \textbf{Acc. Std} & \textbf{Time Mean} & \textbf{Time Std} & \textbf{Epochs Mean} & \textbf{Epochs Std} \\
\midrule
\multirow{7}{*}{CIFAR-10} 
& \multirow{7}{*}{ResNet20} & Adam & - & 0.798 & 0.008 & 63.74 & 3.62 & 28.63 & 1.94 \\
& & AdamW & - & 0.791 & 0.014 & 64.71 & 7.06 & 28.70 & 3.84 \\
& & Adamax & - & 0.689 & 0.015 & 54.89 & 4.43 & 23.73 & 2.45 \\
& & CAdam & Additive & 0.821 & 0.006 & 66.62 & 4.18 & 30.00 & 2.26 \\
& & CAdam & Depth & 0.816 & 0.010 & 66.83 & 5.77 & 30.07 & 3.11 \\
& & CAdam & Multiplicative & \textbf{0.831} & 0.005 & 73.73 & 7.38 & 33.90 & 4.14 \\
& & Nadam & - & 0.792 & 0.014 & 67.79 & 5.89 & 29.00 & 3.13 \\
\midrule
\multirow{7}{*}{CIFAR-100} 
& \multirow{7}{*}{ResNet20} & Adam & - & 0.478 & 0.006 & 71.83 & 3.98 & 32.50 & 2.16 \\
& & AdamW & - & 0.483 & 0.006 & 72.08 & 3.58 & 32.17 & 1.93 \\
& & Adamax & - & 0.396 & 0.009 & 79.36 & 3.37 & 36.57 & 1.76 \\
& & CAdam & Additive & 0.491 & 0.010 & 70.63 & 3.56 & 31.80 & 1.86 \\
& & CAdam & Depth & 0.504 & 0.005 & 73.12 & 4.33 & 32.90 & 2.37 \\
& & CAdam & Multiplicative & \textbf{0.507} & 0.005 & 69.29 & 3.56 & 30.93 & 1.93 \\
& & Nadam & - & 0.476 & 0.011 & 74.23 & 4.23 & 31.77 & 2.18 \\
\midrule
\multirow{7}{*}{Fashion-MNIST} 
& \multirow{7}{*}{ResNet20} & Adam & - & 0.923 & 0.005 & 139.01 & 12.82 & 32.00 & 3.27 \\
& & AdamW & - & 0.924 & 0.004 & 140.31 & 16.86 & 32.13 & 4.23 \\
& & Adamax & - & 0.916 & 0.004 & 128.91 & 12.88 & 29.50 & 3.26 \\
& & CAdam & Additive & 0.925 & 0.004 & 150.64 & 24.09 & 34.93 & 6.08 \\
& & CAdam & Depth & \textbf{0.926} & 0.003 & 143.59 & 19.05 & 33.13 & 4.80 \\
& & CAdam & Multiplicative & 0.922 & 0.003 & 166.07 & 24.34 & 38.87 & 6.20 \\
& & Nadam & - & 0.925 & 0.003 & 133.95 & 14.71 & 29.80 & 3.70 \\
\bottomrule
\end{tabular}
}
\label{table:classification_raw_results}
\end{table*}

\begin{table*}[!h]
\centering
\caption{Statistical Analysis of RMSE Improvements on California Housing Dataset}
\label{tab:california_statistical_results}
\resizebox{0.8\textwidth}{!}{%
\begin{tabular}{llllrrrrrrr}
\toprule
\textbf{Architecture} & \textbf{Optimizer} & \textbf{Scaling Strategy} & \textbf{Metric} & \multicolumn{3}{c}{\textbf{RMSE}} & \multicolumn{3}{c}{\textbf{Training Time}} \\
\cmidrule(lr){5-7} \cmidrule(lr){8-10}
& & & & \textbf{Improv. (\%)} & \textbf{t-stat} & \textbf{p-value} & \textbf{Improv. (\%)} & \textbf{t-stat} & \textbf{p-value} \\
\midrule
\multirow{6}{*}{[1024, 256, 64, 16]} 
& AdamW & - & RMSE & 0.00 & 0.03 & 9.77e-01 & 0.10 & 0.03 & 9.75e-01 \\
& Adamax & - & RMSE & -2.23 & -18.40 & 6.95e-26\textsuperscript{***} & -26.29 & -5.41 & 1.24e-06\textsuperscript{***} \\
& CAdam & Additive & RMSE & 0.49 & 3.24 & 2.01e-03\textsuperscript{**} & 4.11 & 1.49 & 1.43e-01 \\
& CAdam & Depth & RMSE & \textbf{0.50} & 3.36 & 1.39e-03\textsuperscript{**} & 2.37 & 0.80 & 4.29e-01 \\
& CAdam & Multiplicative & RMSE & -0.26 & -1.40 & 1.67e-01 & 5.15 & 2.04 & 4.62e-02\textsuperscript{*} \\
& Nadam & - & RMSE & 0.22 & 1.45 & 1.52e-01 & -6.80 & -1.66 & 1.01e-01 \\
\midrule
\multirow{6}{*}{[128, 128, 128, 128, 128]} 
& AdamW & - & RMSE & 0.25 & 1.35 & 1.83e-01 & 1.42 & 0.64 & 5.26e-01 \\
& Adamax & - & RMSE & -2.06 & -13.41 & 1.99e-19\textsuperscript{***} & -18.96 & -6.87 & 4.75e-09\textsuperscript{***} \\
& CAdam & Additive & RMSE & 1.07 & 4.80 & 1.17e-05\textsuperscript{***} & 3.56 & 1.54 & 1.30e-01 \\
& CAdam & Depth & RMSE & 1.08 & 5.91 & 1.95e-07\textsuperscript{***} & 0.28 & 0.12 & 9.01e-01 \\
& CAdam & Multiplicative & RMSE & \textbf{1.44} & 7.48 & 4.65e-10\textsuperscript{***} & -2.93 & -1.19 & 2.37e-01 \\
& Nadam & - & RMSE & 0.09 & 0.53 & 6.01e-01 & -6.14 & -2.35 & 2.24e-02\textsuperscript{*} \\
\midrule
\multirow{6}{*}{[128, 128, 64]} 
& AdamW & - & RMSE & 0.06 & 0.65 & 5.21e-01 & 0.76 & 0.32 & 7.51e-01 \\
& Adamax & - & RMSE & -1.82 & -19.20 & 8.28e-27\textsuperscript{***} & -34.88 & -8.51 & 8.46e-12\textsuperscript{***} \\
& CAdam & Additive & RMSE & 1.14 & 9.03 & 1.18e-12\textsuperscript{***} & 8.48 & 3.77 & 3.87e-04\textsuperscript{***} \\
& CAdam & Depth & RMSE & 0.99 & 9.02 & 1.22e-12\textsuperscript{***} & 4.06 & 1.80 & 7.66e-02 \\
& CAdam & Multiplicative & RMSE & \textbf{2.24} & 15.94 & 7.27e-23\textsuperscript{***} & 7.69 & 3.81 & 3.37e-04\textsuperscript{***} \\
& Nadam & - & RMSE & 0.08 & 0.86 & 3.93e-01 & -1.21 & -0.47 & 6.37e-01 \\
\midrule
\multirow{6}{*}{[128, 64, 32]} 
& AdamW & - & RMSE & 0.18 & 1.23 & 2.25e-01 & 0.98 & 0.18 & 8.57e-01 \\
& Adamax & - & RMSE & -1.68 & -12.74 & 1.85e-18\textsuperscript{***} & -61.14 & -3.29 & 1.70e-03\textsuperscript{**} \\
& CAdam & Additive & RMSE & 1.77 & 11.77 & 5.20e-17\textsuperscript{***} & 9.70 & 1.82 & 7.33e-02 \\
& CAdam & Depth & RMSE & 1.26 & 9.80 & 6.55e-14\textsuperscript{***} & 0.35 & 0.06 & 9.51e-01 \\
& CAdam & Multiplicative & RMSE & \textbf{2.87} & 16.89 & 4.60e-24\textsuperscript{***} & 15.41 & 3.09 & 3.06e-03\textsuperscript{**} \\
& Nadam & - & RMSE & -0.17 & -1.27 & 2.11e-01 & 1.54 & 0.28 & 7.81e-01 \\
\midrule
\multirow{6}{*}{[256, 128, 64, 32]} 
& AdamW & - & RMSE & 0.27 & 2.19 & 3.22e-02\textsuperscript{*} & -1.02 & -0.40 & 6.93e-01 \\
& Adamax & - & RMSE & -1.52 & -15.59 & 2.12e-22\textsuperscript{***} & -32.52 & -6.59 & 1.45e-08\textsuperscript{***} \\
& CAdam & Additive & RMSE & 1.15 & 6.80 & 6.24e-09\textsuperscript{***} & 4.99 & 1.74 & 8.68e-02 \\
& CAdam & Depth & RMSE & 0.93 & 6.67 & 1.05e-08\textsuperscript{***} & 0.41 & 0.15 & 8.84e-01 \\
& CAdam & Multiplicative & RMSE & \textbf{1.43} & 8.53 & 7.90e-12\textsuperscript{***} & -0.14 & -0.05 & 9.63e-01 \\
& Nadam & - & RMSE & -0.30 & -2.57 & 1.28e-02\textsuperscript{*} & -3.25 & -1.12 & 2.69e-01 \\
\midrule
\multirow{6}{*}{[64, 32]} 
& AdamW & - & RMSE & 0.06 & 0.38 & 7.05e-01 & 11.48 & 1.09 & 2.80e-01 \\
& Adamax & - & RMSE & -1.24 & -8.60 & 6.04e-12\textsuperscript{***} & -79.17 & -3.69 & 4.96e-04\textsuperscript{***} \\
& CAdam & Additive & RMSE & 0.61 & 3.53 & 8.23e-04\textsuperscript{***} & 25.90 & 3.00 & 3.96e-03\textsuperscript{**} \\
& CAdam & Depth & RMSE & 0.41 & 2.23 & 2.97e-02\textsuperscript{*} & 16.79 & 1.69 & 9.72e-02 \\
& CAdam & Multiplicative & RMSE & \textbf{2.07} & 13.30 & 2.89e-19\textsuperscript{***} & 30.11 & 3.12 & 2.79e-03\textsuperscript{**} \\
& Nadam & - & RMSE & 0.31 & 1.93 & 5.86e-02 & 18.19 & 2.01 & 4.92e-02\textsuperscript{*} \\
\bottomrule
\end{tabular}
}
\label{table:california_statistical_results}
\end{table*}

\begin{table*}[!h]
\centering
\caption{Statistical Analysis of Performance Improvements Compared to Adam Baseline}
\label{tab:statistical_results}
\resizebox{0.8\textwidth}{!}{%
\begin{tabular}{llllrrrrrrr}
\toprule
\textbf{Dataset} & \textbf{Optimizer} & \textbf{Scaling Strategy} & \textbf{Metric} & \multicolumn{3}{c}{\textbf{Accuracy}} & \multicolumn{3}{c}{\textbf{Training Time}} \\
\cmidrule(lr){5-7} \cmidrule(lr){8-10}
& & & & \textbf{Improv. (\%)} & \textbf{t-stat} & \textbf{p-value} & \textbf{Improv. (\%)} & \textbf{t-stat} & \textbf{p-value} \\
\midrule
\multirow{6}{*}{CIFAR-10} 
& AdamW & - & Accuracy & -0.93 & -2.49 & 1.58e-02 & -1.51 & -0.67 & 5.08e-01 \\
& Adamax & - & Accuracy & -13.67 & -35.53 & 4.62e-41\textsuperscript{***} & 13.88 & 8.47 & 1.02e-11\textsuperscript{***} \\
& CAdam & Additive & Accuracy & 2.84 & 12.70 & 2.16e-18\textsuperscript{***} & -4.51 & -2.85 & 6.08e-03\textsuperscript{**} \\
& CAdam & Depth & Accuracy & 2.20 & 7.56 & 3.32e-10\textsuperscript{***} & -4.84 & -2.48 & 1.59e-02\textsuperscript{*} \\
& CAdam & Multiplicative & Accuracy & \textbf{4.09} & 18.89 & 1.89e-26\textsuperscript{***} & -15.67 & -6.66 & 1.09e-08\textsuperscript{***} \\
& Nadam & - & Accuracy & -0.82 & -2.21 & 3.08e-02\textsuperscript{*} & -6.35 & -3.21 & 2.18e-03\textsuperscript{**} \\
\midrule
\multirow{6}{*}{CIFAR-100} 
& AdamW & - & Accuracy & 1.02 & 3.19 & 2.29e-03\textsuperscript{**} & -0.34 & -0.25 & 8.01e-01 \\
& Adamax & - & Accuracy & -17.21 & -39.92 & 6.99e-44\textsuperscript{***} & -10.49 & -7.92 & 8.49e-11\textsuperscript{***} \\
& CAdam & Additive & Accuracy & 2.72 & 6.04 & 1.17e-07\textsuperscript{***} & 1.67 & 1.23 & 2.25e-01 \\
& CAdam & Depth & Accuracy & 5.48 & 17.32 & 1.37e-24\textsuperscript{***} & -1.80 & -1.20 & 2.34e-01 \\
& CAdam & Multiplicative & Accuracy & \textbf{5.97} & 20.00 & 1.06e-27\textsuperscript{***} & 3.53 & 2.60 & 1.17e-02\textsuperscript{*} \\
& Nadam & - & Accuracy & -0.39 & -0.82 & 4.18e-01 & -3.35 & -2.27 & 2.72e-02\textsuperscript{*} \\
\midrule
\multirow{6}{*}{Fashion-MNIST} 
& AdamW & - & Accuracy & 0.15 & 1.24 & 2.19e-01 & -0.93 & -0.34 & 7.39e-01 \\
& Adamax & - & Accuracy & -0.76 & -6.36 & 3.42e-08\textsuperscript{***} & 7.26 & 3.04 & 3.51e-03\textsuperscript{**} \\
& CAdam & Additive & Accuracy & 0.31 & 2.66 & 1.00e-02\textsuperscript{*} & -8.36 & -2.33 & 2.31e-02\textsuperscript{*} \\
& CAdam & Depth & Accuracy & \textbf{0.39} & 3.61 & 6.35e-04\textsuperscript{***} & -3.30 & -1.09 & 2.79e-01 \\
& CAdam & Multiplicative & Accuracy & -0.02 & -0.23 & 8.22e-01 & -19.47 & -5.39 & 1.00e-06\textsuperscript{***} \\
& Nadam & - & Accuracy & 0.23 & 2.16 & 3.48e-02\textsuperscript{*} & 3.64 & 1.42 & 1.61e-01 \\
\bottomrule
\end{tabular}
}
\label{table:classification_statistical_results}
\end{table*}

\end{document}